\documentclass{article}

\PassOptionsToPackage{numbers}{natbib}
\usepackage[dblblindworkshop, final]{neurips_2025}
\workshoptitle{Learning to Sense}

\usepackage[utf8]{inputenc} % allow utf-8 input
\usepackage[T1]{fontenc}    % use 8-bit T1 fonts
\PassOptionsToPackage{hyphens}{url}
\usepackage{booktabs}       % professional-quality tables
\usepackage{amsfonts}       % blackboard math symbols
\usepackage{nicefrac}       % compact symbols for 1/2, etc.
\usepackage{microtype}      % microtypography
\usepackage[dvipsnames]{xcolor}
\usepackage{amsmath}

\usepackage{graphicx}
\usepackage{subcaption}
\usepackage{microtype} % helps with tight spacing
\usepackage[colorlinks, citecolor=Blue, linkcolor=Blue, urlcolor=Blue, backref=page]{hyperref}

\definecolor{maroon}{RGB}{128,0,0}

\title{Inference-Time Scaling of Diffusion Models\\for Infrared Data Generation}

\author{%
  Kai A. Horstmann\thanks{Correspondence: \texttt{kah288@cornell.edu}; work done while an intern at YRIKKA.} \\
  Cornell University \\
  \And
  Maxim Clouser \\
  YRIKKA, Inc. \\
  \And
  Kia Khezeli \\
  YRIKKA, Inc. \\
}

\begin{document}

\maketitle

\setcounter{footnote}{0}

%%%%%%%%%%%%%%%%% BEGIN CONTENT %%%%%%%%%%%%%%%%%

\begin{abstract}
Infrared imagery enables temperature-based scene understanding using passive sensors, particularly under conditions of low visibility where traditional RGB imaging fails. Yet, developing downstream vision models for infrared applications is hindered by the scarcity of high-quality annotated data, due to the specialized expertise required for infrared annotation.
While synthetic infrared image generation has the potential to accelerate model development by providing large-scale, diverse training data, training foundation-level generative diffusion models in the infrared domain has remained elusive due to limited datasets.
In light of such data constraints, we explore an inference-time scaling approach using a domain-adapted CLIP-based verifier for enhanced infrared image generation quality.
We adapt FLUX.1-dev, a state-of-the-art text-to-image diffusion model, to the infrared domain by finetuning it on a small sample of infrared images using parameter-efficient techniques.
The trained verifier is then employed during inference to guide the diffusion sampling process toward higher quality infrared generations that better align with input text prompts.
Empirically, we find that our approach leads to consistent improvements in generation quality, reducing FID scores on the KAIST Multispectral Pedestrian Detection Benchmark dataset by 10\% compared to unguided baseline samples. Our results suggest that inference-time guidance offers a promising direction for bridging the domain gap in low-data infrared settings.\end{abstract}

\section{Introduction}\label{sec:intro}

Infrared sensors capture thermal information unavailable through conventional RGB cameras, making them essential for applications ranging from autonomous driving to medical imaging. The growing deployment of computer vision systems in such domains has created demand for large-scale synthetic infrared datasets to train robust downstream models such as object detectors and scene classifiers. Diffusion models \citep{DenoisingDiffusionProbabilisticho2020, ScoreBasedGenerativeModelingsong2021} have emerged as a promising approach for generating such synthetic data, offering the potential to create diverse, high-quality images that can augment limited real-world datasets.

While the ubiquity of color images allows large-scale diffusion models to be pretrained on billions of RGB images sourced from the web, infrared images remain far less accessible and are typically confined to a limited number of curated public datasets.
 This scarcity poses significant challenges for training foundation-level infrared diffusion models.
 Instead, finetuning pretrained RGB models leverages the rich world knowledge encoded from billions of natural images while adapting representations to the infrared modality.
 To this end, prior works \citep{DiffV2IRVisibletoInfraredDiffusionran2025, V2IRCnLDMGenerativeVisibletoinfraredreinhardt2025} have explored techniques to finetune pretrained RGB diffusion models on available public infrared image datasets, albeit with limited physical realism.
 Such approaches suffer from a lack of contextual grounding; while generated outputs may appear thermodynamically plausible, they often fail to remain consistent with the semantic or contextual cues provided by the conditioning input, whether a textual prompt or RGB image.
 Other approaches such as PID \citep{PIDPhysicsInformedDiffusionmao2025} train diffusion models directly on limited infrared data with physics-informed constraints, but exhibit inconsistent generation quality and mode collapse, highlighting the pitfalls of direct training in the absence of large-scale pretraining. 

To address such challenges, we turn to inference-time scaling, an extension of training-time neural scaling laws \citep{ScalingLawsNeuralkaplan} commonly observed during the training of deep learning models. Inference-time scaling methods aim to improve the performance of models post hoc by expending additional compute during inference, and have shown promise in their recent application to large language models \citep{ChainofThoughtPromptingElicitswei2023, TreeThoughtsDeliberateyao2023, ScalingLLMTestTimesnell2024} and diffusion models \citep{InferenceTimeScalingDiffusionma2025, TestTimeScalingDiffusionramesh2025, GeneralFrameworkInferencetimesinghal2025}.
Motivated by these advances, we investigate the efficacy of inference-time scaling in improving the quality of synthetic infrared images generated by diffusion models, a direction which, to the best of our knowledge, has not previously been explored.
Whereas large-scale pretraining is constrained by infrared data scarcity, we now shift the computational burden from training to generation, thereby improving the visual quality of infrared images produced by diffusion models finetuned on limited data.
Building on the inference-time scaling framework of \citet{InferenceTimeScalingDiffusionma2025} which combines sampling algorithms with verifiers, we extend this approach to the infrared domain and introduce a novel self-supervised verifier designed to evaluate the consistency and realism of synthetic infrared images.
In particular, we finetune CLIP \citep{LearningTransferableVisualradford2021} to distinguish true infrared images from their grayscale counterparts, and leverage this model as an inference-time verifier to guide iterative search through the noise space, selecting noise latents that yield the highest-quality generations. Our experiments demonstrate measurable improvements in FID scores, and we further assess how different noise search strategies and verifier training protocols influence inference-time scaling performance.

\newcommand{\imagescore}[5]{
  \includegraphics[width=\linewidth]{#1}\par
  \vspace{2pt}
  {\scriptsize
  \centering
  \begin{tabular}{@{}l r@{}}
    \textsc{IRScore} ($\times 10$) (ours):   & #2 \\
    IR Similarity:   & #3 \\
    Grayscale Similarity:   & #4 \\ [1pt] \hline \\[-6.5pt]
    \textsc{IRScore} ($\times 10$) (pretrained): & #5
  \end{tabular}\par}
}

\begin{figure}[t]
  \centering
    \begin{subfigure}[t]{0.32\linewidth}
    \imagescore{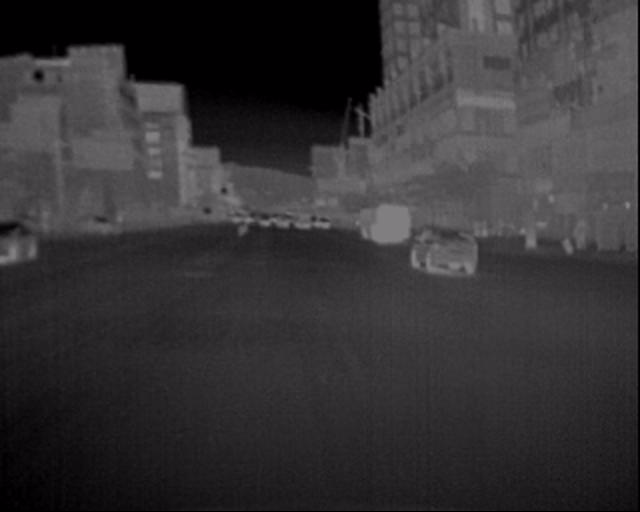}{0.5062}{0.4879}{0.3866}{0.1016}
    \caption{Ground Truth IR}
    \label{fig:true_ir}
  \end{subfigure}\hfill
  \begin{subfigure}[t]{0.32\linewidth}
    \imagescore{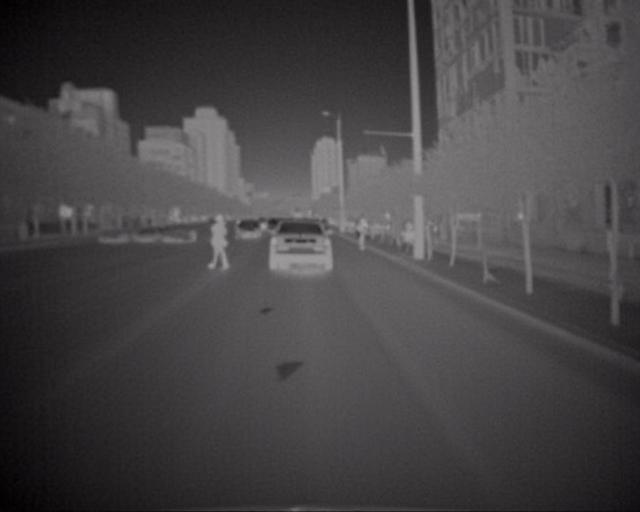}{0.5135}{0.4213}{0.3186}{-0.0414}
    \caption{High-Scoring Synthetic IR}
    \label{fig:good_ir}
  \end{subfigure}\hfill
    \begin{subfigure}[t]{0.32\linewidth}
    \imagescore{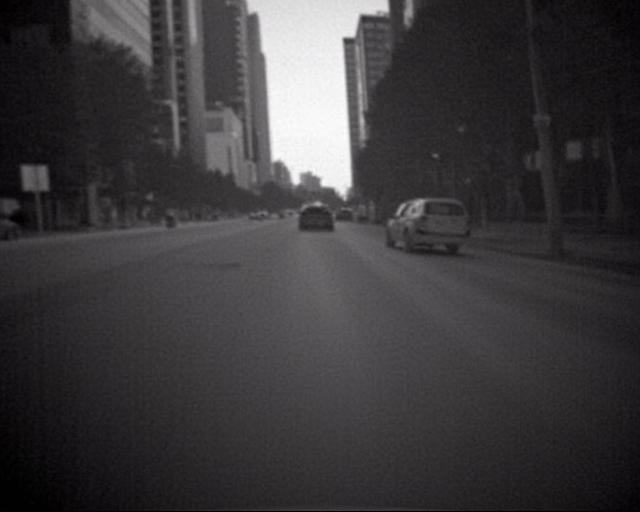}{-0.3212}{0.3189}{0.3831}{-0.0125}
    \caption{Low-Scoring Synthetic IR}
    \label{fig:bad_ir}
  \end{subfigure}\hfill
  \caption{Example verifier scores for a ground truth infrared image and two synthetic images (generated from different random seeds) corresponding to the caption, ``\texttt{A city street at dusk features tall buildings with illuminated signs, a marked road with directional arrows, and vehicles including a white SUV driving away from the camera.}'' \textsc{IRScore} (ours), IR Similarity, and Grayscale Similarity are computed using our finetuned CLIP model, while \textsc{IRScore} (pretrained) relies on a pretrained CLIP model.
  \textsc{IRScore}, IR Similarity, and Grayscale Similarity correspond to Equation~\ref{eq:ir-score}, and its unscaled first and second terms, respectively.
  }
  \label{fig:ir_three_row}
  \vspace{-2mm}
\end{figure}

\section{Background}\label{sec:background}

A naive approach to scaling diffusion models at inference time is to increase the number of denoising steps during image generation. Scaling up denoising steps can improve the visual quality of generated images, albeit with diminishing returns \citep{ElucidatingDesignSpacekarras2022}. Recent works \citep{InferenceTimeScalingDiffusionma2025, TestTimeScalingDiffusionramesh2025, GeneralFrameworkInferencetimesinghal2025} have investigated alternative methods to scaling, motivated by the observation that different noise latents result in varied levels of quality in generated images \citep{NoiseWorthDiffusionahn2024, NotAllNoisesqi2024}. \citet{InferenceTimeScalingDiffusionma2025} extends this finding by introducing a scaling approach consisting of two components: verifiers and search algorithms.

\subsection{Verifiers}\label{ssec:verfiers} 
Concretely, a verifier is defined as a function $\mathcal V: \mathbb{R}^{h \times w} \times \mathbb R^d \to \mathbb R$ which takes in a sample image and an optional condition (e.g., a prompt or image class) and produces a scalar value reflecting the quality of the image, either unconditionally or relative to the provided condition. In the text-to-image setting, \citet{InferenceTimeScalingDiffusionma2025} explores numerous pretrained models as verifiers, each designed to evaluate different notions of image quality. CLIPScore \citep{CLIPScoreReferencefreeEvaluationhessel2021} is chosen for measuring image-prompt alignment, Aesthetic Score Predictor \citep{LAION5BOpenLargescaleschuhmann2022} for human aesthetic preferences, and ImageReward \citep{ImageRewardLearningEvaluatingxu2023} for a more comprehensive metric incorporating both aesthetic quality and prompt adherence. Additionally, vision language models (VLMs) such as Gemini-1.5 Flash are employed for their sophisticated text-image understanding and ability to assess generated images across diverse criteria.

\subsection{Search Algorithms}\label{ssec:search} 
Formally, a search algorithm is some function 
 $f : (\mathcal V, D_\theta, \{ \mathbf z_1, \dots , \mathbf z_N \}) \mapsto \mathbf z^\ast $
which takes as input a verifier $\mathcal V$, a generative diffusion model $D_\theta$, and $N$ candidate noise latents. It then outputs the ``best'' noise latent $\mathbf z^\ast$ corresponding to the highest scoring image among $N$ images sampled from $D_\theta$, as determined by $\mathcal V$. Invoking $f$ requires numerous forward passes through $D_\theta$ which constitutes the primary computational bottleneck; thus, this cost is quantified as the number of function evaluations (NFEs). Following \citet{InferenceTimeScalingDiffusionma2025}, we consider two primary search strategies below.

\paragraph{Random Search.} In random search, $N$ candidate noise latents are sampled from a Gaussian distribution and progressively denoised over a fixed number of steps.
A given verifier evaluates the resulting images, and the highest-scoring image among the $N$ initial latents is returned. The scaling axis in random search is simply $N$; that is, $\text{NFEs} = N \cdot (\#\text{ of denoising steps})$. Notably, random search can lead to \textit{verifier hacking}, a failure mode in which the search overfits to biases of the pretrained verifier rather than producing genuinely higher-quality generations \citep{InferenceTimeScalingDiffusionma2025}.

\paragraph{Zero-Order Search.} To mitigate verifier hacking, zero-order search  traverses the noise space incrementally in directions that are locally optimal with respect to the verifier’s score. Starting from an initial noise latent as the pivot, the algorithm samples $N-1$ additional latents in its neighborhood. All $N$ latents are denoised, and the latent yielding the highest-scoring image becomes the new pivot. This process is repeated for $k$ iterations, resulting in $\text{NFEs} = kN \cdot (\#\text{ of denoising steps})$.

\section{Method}\label{sec:method}

In this work, we focus on the text-to-image generation task in the infrared domain, where, given text prompts, we aim to generate physically plausible infrared images with a pretrained text-to-image diffusion model. To that end, we apply inference-time scaling to guide the generation process toward higher-quality outputs that better reflect infrared imaging characteristics.

\paragraph{Infrared Image Generation.} We select FLUX.1-dev \citep{BlackforestlabsFlux2025}, a state-of-the-art text-to-image model, as our diffusion model backbone. To endow the model with knowledge of the infrared domain, we first apply Low-Rank Adaptation (LoRA) \citep{LoRALowRankAdaptationhu2021} finetuning on a limited set of 1,000 infrared image–caption pairs. We observe that this lightweight adaptation is typically sufficient to equip the model with a basic understanding of infrared image characteristics; however, we find that the resulting generations exhibit high variability in quality, with many samples appearing physically implausible or resembling simple grayscale versions of RGB images (Figure~\ref{fig:ir_three_row}). This inconsistent nature, which we attribute to inductive biases inherited from pretraining on large-scale RGB datasets, creates an opportunity to apply verifier-based inference-time selection to identify and promote higher-quality outputs.

\paragraph{Verifier Training.} To steer generations towards more physically accurate infrared images, we adapt CLIP \citep{LearningTransferableVisualradford2021} as a zero-shot verifier. Since CLIP is pretrained on large-scale RGB image data, it is not suitable for evaluating the quality of infrared images. Accordingly, we finetune CLIP on paired infrared images and text captions, enabling the model to capture the unique characteristics of the infrared domain.
 
 Motivated by our earlier observation that our diffusion model frequently generates grayscale-like images, we augment each training mini-batch with grayscale versions of the corresponding RGB images. Further, we format the text captions as ``\texttt{An INFRARED photo of $\{$caption$\}$.}'' and ``\texttt{A GRAYSCALE photo of $\{$caption$\}$.}'' as appropriate. In doing so, we encourage the model to learn distinct representations for true infrared images and their grayscale counterparts.
 
 \paragraph{Inference-Time Verification.} Let $\Phi_I: \mathbb R^{h \times w} \to \mathbb R^d$ and $\Phi_T: V^{\leq n} \to \mathbb R^d$ denote the finetuned image and text encoders of the CLIP verifier, respectively. At inference time, given a candidate image $\mathbf x \in \mathbb{R}^{h \times w}$ generated by our finetuned diffusion model, along with its corresponding infrared and grayscale captions $c_\text{IR}$ and $c_\text{gray}$, we compute an infrared quality score as
 \begin{equation}\label{eq:ir-score}
 	\operatorname{\textsc{IRScore}}(\mathbf x) =
 	(1-\alpha) \cos(\Phi_T(c_\text{IR}), \Phi_I(\mathbf x)) - \alpha \cos(\Phi_T(c_\text{gray}), \Phi_I(\mathbf x)),
 \end{equation}
where $\alpha \in [0, 1]$ is a tunable hyperparameter that  balances alignment with true infrared semantics against undesired similarity to grayscale. This verifier-driven score serves as a selection criterion during sampling; we evaluate multiple candidate generations and retain those that maximize the score using either a random or zero-order search strategy, as described in Section~\ref{ssec:search}.

\begin{table}
  \caption{Performance of two inference-time scaling methods on KAIST dataset.}
  \label{tab:main-results}
  \centering
\begin{tabular}{llllll}
\toprule
Method & NFEs & $\alpha$ & Split & \textsc{IRScore} ($\times 10$) $\uparrow$ & FID $\downarrow$ \\ 
\midrule
  Naive Sampling & 28 & 0.5 & Test & 0.1187 & 74.58 \\
  \midrule
  Random Search & 336 & 0.5 & Test & \textbf{0.6010} & \textbf{66.74} \\
  Zero-Order Search & 336 & 0.5 & Test & 0.4214 & 69.15 \\
  \bottomrule
\end{tabular}
\end{table}

\section{Experiments}\label{sec:experiments}

\paragraph{Setup.} We conduct experiments on 49,561 images\footnote{\url{https://huggingface.co/datasets/koifisharriet/KAIST-Multispectral-Pedestrian-Benchmark}} from the KAIST Multispectral Pedestrian Detection Benchmark \citep{MultispectralPedestrianDetectionhwang2015}, a dataset of long-wave infrared and RGB pedestrian scenes captured from a vehicle. The data is divided into an 80/20 train–test split. As our diffusion backbone, we adopt FLUX.1-dev, a pretrained text-to-image model, augmented with a LoRA adapter of rank $r=16$ finetuned on 1,000 infrared images from the KAIST training set along with corresponding captions. For verification, we employ CLIP-B/32 \citep{LearningTransferableVisualradford2021}, finetuned on the same KAIST training data following the strategy outlined in Section~\ref{sec:method}.

 \paragraph{Results.} Our experiments suggest that inference-time scaling can result in meaningful gains in infrared generative quality. In particular, we observe a reduction in the Fréchet Inception Distance (FID) \citep{GANsTrainedTwoheusel2017} score from 74.58 to 66.74 on the KAIST dataset (Table~\ref{tab:main-results}). The FID quantifies the discrepancy between the distribution of generated images and that of real images, with lower values indicating closer alignment; thus, this improvement provides promising evidence that inference-time scaling can enhance generative performance. Between the two search algorithms explored, random search achieves the largest improvement, reducing FID by over 10\% relative to the naive baseline, while zero-order search achieves a more modest $\sim$7\% gain under the same NFE budget. This aligns with the findings of \citet{InferenceTimeScalingDiffusionma2025} that the incremental update procedure of zero-order search results in slower convergence, as opposed to random search, which enables a broader exploration of the noise space. We also note that our NFE budget of 336 is relatively limited compared to other works that have explored inference-time scaling \citep{InferenceTimeScalingDiffusionma2025, TestTimeScalingDiffusionramesh2025}, and we anticipate that performance could improve substantially by further scaling up the computational budget.
 
 Further, we find qualitatively that our domain-adapted CLIP is effective in distinguishing between high and low-quality synthetic infrared images. As illustrated in Figure~\ref{fig:ir_three_row}, the verifier assigns substantially higher scores to realistic infrared images that exhibit proper thermal characteristics, whereas an unmodified CLIP model fails to properly distinguish between such quality differences, instead assigning a higher score to the lower-quality image. We also report the mean of the \textsc{IRScore} (Equation~\ref{eq:ir-score}) across all generated samples in Table~\ref{tab:main-results}, and find that both random and zero-order search effectively explore the noise space to identify latents with higher verifier scores, yielding consistent improvements over naive sampling.

\section{Conclusion}
In this work, we introduced an inference-time scaling approach using a domain-adapted CLIP verifier for enhanced infrared image generation quality in text-to-image diffusion models. Our key contribution is training CLIP to distinguish between realistic infrared images and grayscale-like artifacts, enabling effective guidance of diffusion model sampling. Preliminary results on the KAIST dataset demonstrate FID score improvements of over 10\% compared to unguided baselines, establishing inference-time guidance as a promising direction for infrared image generation in low-data settings.
Building on this foundation, promising avenues for future research include training alternative verifiers such as physics-based or domain-specific models, as well as extending evaluation to other datasets and sensing modalities to assess the generality of the approach.

\bibliographystyle{unsrtnat}
\bibliography{clip_ir}

%%%%%%%%%%%%%%%%% END CONTENT %%%%%%%%%%%%%%%%%

\end{document}